\documentclass{article}

\usepackage{microtype}
\usepackage{graphicx}
\usepackage{subfigure}
\usepackage{booktabs}

\usepackage{hyperref}
\usepackage{color}

\usepackage[accepted]{icml2020}

\usepackage{amsmath, amsfonts, amssymb, amsthm, bm}

\usepackage{amsmath,amssymb}
\usepackage{graphicx}
\usepackage{color}
\usepackage{bm}
\usepackage{algorithm}
\usepackage{algorithmic}
\usepackage{wrapfig}
\usepackage{hyperref}
\usepackage{url}

\newtheorem{theorem}{Theorem}
\newtheorem{proposition}[theorem]{Proposition}

\newcommand{\bbox}{\hfill $\Box$}

\newcommand{\T}{ {\mathrm{\scriptscriptstyle T}} }

\newcommand{\benr}{\begin{eqnarray}}
\newcommand{\eenr}{\end{eqnarray}}
\newcommand{\benrr}{\begin{eqnarray*}}
\newcommand{\eenrr}{\end{eqnarray*}}
\newcommand{\ben}{\begin{equation}}
\newcommand{\een}{\end{equation}}
\newcommand{\benn}{\begin{equation*}}
\newcommand{\eenn}{\end{equation*}}

\newcommand{\nn}{\nonumber}

\newcommand{\cE}{\mathcal E}
\newcommand{\cF}{\mathcal F}
\newcommand{\cH}{\mathcal H}
\newcommand{\cL}{\mathcal L}
\newcommand{\cP}{\mathcal P}
\newcommand{\cQ}{\mathcal Q}

\newcommand{\cX}{\mathcal X}
\newcommand{\cY}{\mathcal Y}

\def\vone{\bm{1}}

\def\vr{\bm{r}}
\def\vq{\bm{q}}

\def\vv{\bm{v}}

\def\mI{{\bm{I}}}

\def\rr{{\textnormal{r}}}

\def\ry{{\textnormal{y}}}

\def\rvr{{\mathbf{r}}}

\def\rvx{{\mathbf{x}}}
\def\rvy{{\mathbf{y}}}
\def\rvz{{\mathbf{z}}}

\newcommand{\E}{\mathbb{E}}
\newcommand{\Var}{\mathrm{Var}}

\newcommand{\R}{\mathbb R}

\DeclareMathOperator*{\argmax}{arg\,max}
\DeclareMathOperator*{\argmin}{arg\,min}

\newcommand{\rarrow}{\rightarrow}

\def\sD{{\mathbb{D}}}

\def\sP{{\mathbb{P}}}

\def\sS{{\mathbb{S}}}

\icmltitlerunning{Risk Variance Penalization}

\begin{document}
	
\twocolumn[
\icmltitle{Risk Variance Penalization}

\icmlsetsymbol{equal}{1}
\icmlsetsymbol{corresponding}{*}

\begin{icmlauthorlist}
	\icmlauthor{Chuanlong Xie}{equal}
	\icmlauthor{Haotian Ye}{equal}
	\icmlauthor{Fei Chen}{}
	\icmlauthor{Yue Liu}{}
	\icmlauthor{Rui Sun}{}
	\icmlauthor{Zhenguo Li}{corresponding}
\end{icmlauthorlist}



\icmlkeywords{Out-of-Distribution, Domain Generalization, Risk Variance Penalization}

\vskip 0.3in
]


\begin{abstract}

The key of the out-of-distribution (OOD) generalization is to generalize invariance from training domains to target domains. The variance risk extrapolation (V-REx, \citet{krueger2020out}) is a practical OOD method, which depends on a domain-level regularization but lacks theoretical verifications about its motivation and function. This article provides theoretical insights into V-REx by studying a variance-based regularizer. We propose Risk Variance Penalization (RVP), which slightly changes the regularization of V-REx but addresses the theory concerns about V-REx.
We provide theoretical explanations and a theory-inspired tuning scheme for the regularization parameter of RVP. Our results point out that RVP discovers a robust predictor. Finally, we experimentally show that the proposed regularizer can find an invariant predictor under certain conditions.


\end{abstract}

\section{Introduction}\label{sec:introduction}

The mismatch between training and test data is one major challenge for many machine learning systems, which assume that both training and test data are independent and identically distributed.
However, this assumption does not always hold in practice \citep{bengio2019meta}.
Consequently, its prediction performance
is often degraded when there exist distribution shifts. Two common examples are the problem of identifying cows and camels in different backgrounds \citep{beery2018recognition} and the classification task on the ColoredMNIST dataset \citep{arjovsky2019invariant}.

The Out-of-Distribution (OOD) generalization
is a rapidly growing area. 
A typical OOD problem is to find a model with the uniformly good performance over a set of target domains, e.g. research centers, times, experimental conditions and so on.
Suppose the training data $\{(\rvx^e, \rvy^e)\}$ is multi-domain sourced with known domain label. 
Let $\cE_{tr}$ be the set of the training domains that is a subset of all target domains $\cE.$
The problem is to learn a predictor based on data from $\cE_{tr}$ such that the predictor also works well for any test data from $\cE.$  
In other words, the learned model is robust to the changes over the domain-level.
Notice that some target domains are unseen. 
So this is a typical domain generalization problem that  learns features or structures which guarantee that $P(\rvy^e, \rvx^e)$, $P(\rvy^e|\rvx^e)$ or $\E[\rvy^e|\rvx^e]$ are invariant across all target domains $\cE.$

Intuitively, the training data may inherit some features from $\cE_{tr}$, which varies among $\cE$ but is strongly related to the target $\rvy^e$ under $\cE_{tr}.$ 
In practice, without considering the domain structure, e.g. the shuffle operation, the learner will absorb all the correlations in the pooled training data and learn a model based on all features which are strongly related to the
target \citep{jabri2016revisiting,sturm2014simple,torralba2011unbiased}.  
In the example of classifying cow and camel \citet{beery2018recognition}, cows appear in pastures at most pictures and camels are taken in deserts.
The empirical risk minimization (ERM, \citet{vapnik1992principles}) may learn to recognize cows and camels with background features and
struggle to detect cows in the desert and camels in pastures. This implies that the ERM-learned model may
have a good performance on $\cE_{tr}$, but may dramatically fail under some unseen environments.
However, due to the sampling at the domain-level, the
ERM solution still shows stable performance on some domain generalization tasks \citep{gulrajani2020search}.

It is well known that if we can intervene on the input or change the domains, the generalization performance of a causal model is more stable than that of a
non-causal model.
Thus, some recent works consider bridging the invariance from $\cE_{tr}$ to $\cE$ via causal relationship.
The Invariant Causal Prediction (ICP, \citet{peters2016causal}) uses the invariance under different training domains for causal discovery and inference.
ICP does not estimate graphical, structural equation or potential outcome models but has theoretical guarantees.
\citet{buhlmann2018invariance} unifies some works for causal inference and predictive robustness, e.g. ICP \citep{peters2016causal} and Anchor Regression \citep{rothenhausler2018anchor}.
For large-scale neural networks, the Invariant Risk Minimization (IRM, \citet{arjovsky2019invariant}) is a method suitable for dealing with modern deep learning tasks and can generalize the invariance from $\cE_{tr}$ to $\cE$ under certain assumptions. 
\citet{rosenfeld2020risks} proves that IRM may fail catastrophically unless the test domains are sufficiently similar to the training domains.
On the other hand, \citet{krueger2020out} considers the invariant distribution of $(\rvx^e, \rvy^e)$ and proposes the risk extrapolation method (REx) by extending the group distributional robustness optimization (group DRO, \citet{sagawa2019distributionally}). 
They experimentally show that REx can discover a stable predictor and also deal with the covariate shift since $P(\rvx^e, \rvy^e) = P(\rvx^e) P(\rvy^e|\rvx^e).$ 
In addition to the causal structure and the distributional robustness, there are other definitions of invariance, e.g. conditional independence \citep{koyama2020out}.

\citet{krueger2020out} disentangles the data shift into the shift in $P(\rvy^e|\rvx^e)$ and the shift in $P(\rvx^e)$ (covariate shift).
\citet{koh2020wilds} introduces a meta distribution on domains and denotes the meta distribution shift as "subpopulation shift". 
This paper considers the group distributional robustness which involves these three kinds of distribution shifts.
We focus on two REx methods: Variance REx (V-REx) and Minmax REx (MM-REx).
Although V-REx is doing well on some synthetic and real datasets,it lacks theoretical evidence to support its utility. 
On the other hand, V-REx is a variance-based regularization method.  It does not seem appropriate to use REx to name this approach.
In this paper, we make attempts to answer these questions.
We propose a variance-based regularization method, Risk Variance Penalization (RVP), and prove that RVP is equivalent to MM-REx. 
Besides, the regularization term of RVP is just a tiny change to that of V-REx. That connects V-REx to MM-REx. 
On the other hand, we show that RVP is a quantile regression method and theoretically explain the regularization parameter.  
Furthermore, we propose a theory-inspired tuning scheme.

\section{Related Works}\label{sec: relatedwork}

{\bf Generalization of invariance} The key of OOD problem is to generalize the invariance from training domains to all target domains. Causal inference is a feasible technical route, since  causality leads to invariance.
The understanding that causal variables lead to invariance can be traced back to \citet{haavelmo1943statistical}.
However, the relationship from invariance to causal discovery has not been considered until the Invariant Causal Prediction (ICP, \citep{peters2016causal}). 
It has theoretical guarantees to discover a subset of the causal features without fitting any graphical, structural equation or potential outcome models. 
Thus ICP generalizes the invariance from training domains to all target domains via causal relationship.
More related causal works include \citet{li2017deeper,kuang2018stable,rojas2018invariant,buhlmann2018invariance,magliacane2018domain,hu2018causal,huang2020causal}.
The ICP technique depends on multiple hypothesis testing, which prevents its usage on large scale models. 
\citet{arjovsky2019invariant} proposes a regularization method, Invariant Risk Minimization (IRM), which is compatible with modern deep learning tasks.
\citet{ahuja2020invariant} considers IRM as finding the Nash equilibrium of an ensemble game among several domains and develops a simple training algorithm.
However, the utility of IRM and the tuning scheme has always been controversial, e.g. \citet{rosenfeld2020risks,ahuja2020empirical,gulrajani2020search}.

{\bf Distribution Shift}
The mismatch between training and test data
has been studied as the problem of dataset 
shift \citep{quionero2009dataset}. 
Various forms of dataset shift have been characterized by decomposing data generating distributions, e.g. covariate shift \citep{sugiyama2007covariate,gretton2009covariate},
target shift \citep{zhang2013domain,lipton2018detecting}, conditional shift \citep{zhang2015multi,gong2016domain},  policy shift \citep{schulam2017reliable}, and subpopulation shift \citep{koh2020wilds}.
One class of practical solutions, which can be compatible with large scale models, considers bounded distributional robustness. These methods assume that the test distributions belong to an uncertainty set centred around the training distribution.

{\bf Distributional robustness}
Distributional robustness optimization (DRO) considers a minimax problem over an uncertainty set that is a neighbour set around the truth training distribution or the empirical distribution of training data \citep{ben2013robust,duchi2016statistics,lam2016robust,jiang2016data,lam2017empirical,namkoong2017variance,esfahani2018data,bertsimas2018data,blanchet2019quantifying}.
When the radius of the uncertainty set is small, the objective of DRO can be approximated by a regularized loss \citep{shafieezadeh2015distributionally,namkoong2017variance,duchi2019variance}.
On the other hand, \citet{duchi2019variance} and \citet{lam2017empirical} show the connection between the empirical objective of DRO and confidence bounds of the population objective of ERM via generalized empirical likelihood \citep{owen1990empirical}.
The group Distributional Robustness Optimization considers distribution shifts over groups or domains instead of data points \citep{hu2018does,oren2019distributionally,sagawa2019distributionally}.
The risk extrapolation (REx, \citep{krueger2020out}) is an extension of the group DRO. It introduces 'negative probability' or 'negative density' such that group DRO with large uncertainty set can still be approximated by a regularizer.  


\section{Preliminaries}\label{sec: preliminaries}

Consider a learning task, in which the training data is collected from multiple domains, e.g. research centers, times, experimental conditions and so on.
Suppose that the $n$ training domains $\cE_{tr}$ are randomly selected from all possible domains $\cE$ with a meta distribution $Q$, e.g. 
\benn
\cE_{tr} \subset \cE, \,\, \#\{\cE_{tr}\} = n, \,\, \text{and} \quad e \sim Q, \,\, \forall e \in \cE_{tr}.
\eenn
For each training domain, a set of data points is observed:
\benn
\sD^e = \{\rvz_{1}^e, \ldots, \rvz_{m_e}^e\}, \quad \rvz_{i}^e=(\rvx_{i}^e, \rvy_{i}^e)\sim P^e
\eenn
where $\rvz_{i}^e$ is a data point 
consisting of an input $\rvx_i^e \in \cX$ and the corresponding target $\rvy_i^e \in \cY$, and $P^e$ is the data generating distribution which represents the learning task under the training domain $e.$
Let $\cH$ be the hypothetical space and $h \in \cH$ be a hypothetical model that maps $\rvx \in \cX$ to $h(\rvx) \in \cY.$
The loss function $\ell(\hat y, y): \cY \times \cY \rarrow \R$ measures how poorly the output $\hat y = h(x)$ predicts the target $y.$
If the purpose is to look for the model whose expected 
performance is optimal:
\benrr
h_{erm} &=& \argmin_{h \in \cH} \E_{e \sim Q}\big\{ \E_{\rvz \sim P^e}[ \ell(h(\rvx), \rvy) ]\big\} \\
&=:& \argmin_{h \in \cH} \E_{e \sim Q}\big\{ \rr(h, e) \big\},
\eenrr 
where $\rr(h, e) =  \E_{\rvz \sim P^e}[ \ell(h(\rvx), \rvy) ]$ represents the risk of $h\in \cH$ under the environment $e\in \cE.$ By the sample average approximation, we obtain the objective function of the empirical risk minimization (ERM, \citet{vapnik1992principles}): 
\benrr
\cL_{erm}(h) &=& \frac{1}{\sum_{e \in \cE_{tr}} m_e} \sum_{e \in \cE_{tr}} \sum_{i =1}^{m_e} \ell(h(\rvx_i^e), \rvy_i^e) \\
&=:& \hat \vq_n^\T \hat \rvr_n(h),  
\eenrr
where
\benrr
\hat \vq_n &=& \big[\cdots, \frac{m_e}{\sum_{e \in \cE_{tr}} m_e},
\cdots \big], \\
\hat \rvr_n(h) &=& \big[ \cdots, \hat \rr_m(h, e), \cdots \big] \\
\hat \rr_m(h, e) &=& \frac{1}{m_e} \sum_{i=1}^{m_e} \ell(h(\rvx_i^e), \rvy_i^e).
\eenrr
The group distributional robust optimization (group DRO, \citet{sagawa2019distributionally}) allows us to learn models that minimize the worst-case loss over domains in the training data.
The uncertainty set is any mixture of the training domains, i.e. $\{\sum_e q_e P^e: e \in \cE_{tr}, \vq=(\cdots, q_e, \cdots) \in \Delta_n\}$ where $\Delta_n$ is the $(n-1)$-dimensional probability simplex. 
The objective function of group DRO is given by
\benn
\cL_{dro} (h) = \max_{\vq \in \Delta_n} \vq^\T \hat \rvr_n(h) = \max_{e \in \cE_{tr}} \hat{\rr}_m(h, e)
\eenn
where the second equality holds because the optimum
can be attained at a vertex.
\citet{arjovsky2019invariant} points out that group DRO may use unstable features over training domains and fail to discover an invariant model.

\citet{krueger2020out} extends group DRO by enlarging the uncertainty set and proposes the risk extrapolation method: MM-REx, which assumes the available set of $\vq$ is: 
\benn
\tilde \Delta_n(\alpha):= \{\vq: \vone^\T \vq = 1, \vq + \alpha \vone \in \R^n_{+}\}.
\eenn
Here $\vone$ is a vector of ones. If $\alpha$ is a positive scalar, the elements of $\vq$ can be negative.
The objective function of MM-REx is defined by
\benrr
\cL_{mm-rex} (h) &=& \max_{\vq \in \tilde \Delta_n(\lambda)} \vq^\T \hat \rvr_n(h).
\eenrr
Furthermore, a more practical version, Variance Risk Extrapolation (V-REx), is proposed:
\benrr
\cL_{v-rex}(h) = \frac{1}{n} \vone^\T \hat \rvr_n(h) + \beta s^2_n(\hat \rvr_n(h)) 
\eenrr
where $\beta$ is a tuning parameter and $ s^2_n(\hat \rvr_n(h))$ stands for the sample variance of $\hat \rr_m(h, e)$, i.e.
\benrr
s_n^2(\vr) = \frac{1}{n} \Big\| \vr - \frac{1}{n} \vone \vone^\T \vr \Big\|_2^2.
\eenrr
\citet{krueger2020out} experimentally proves that MM-REx and V-REx can discover the invariant prediction on Colored MNIST \citep{arjovsky2019invariant}. 
However, it remains unclear how MM-REx discovers invariant prediction.
On the other hand, the V-REx method, which  minimizes a variance-regularized objective function, is not derived under the framework of risk extrapolation.
In the following, we shall answer these two questions.

\section{Risk Variance Penalization}\label{sec:Method}

In this section, we propose a variance-based regularization method: Risk Variance Penalization (RVP).
The regularization of RVP is closed to that of V-REx.
Then we connect V-REx to MM-REx by showing that MM-REx is equivalent to RVP. 
On the other hand, we theoretically explain the tuning parameter of RVP and give a theory-inspired tuning scheme.

\subsection{Connect V-REx to MM-REx}\label{sec41}

Let $\phi(x) = (x-1)^2$, which is a convex function $\phi: \R_{+} \rightarrow \R$ with $\phi(1)=0.$ Then the $\phi$-divergence between two distributions $Q$ and $Q'$ is defined by \benn
D_{\phi}(Q|Q') = \int_{\cP} \phi(\frac{dQ}{dQ'}) dQ'.
\eenn
Further, denote $\vq=(q_1, q_2, \ldots, q_n)$ and
\benrr
\cQ_n(\alpha, \rho) &=& \big\{ \vq: \,\, \vq + \alpha \vone \in \R^n_{+}, \,\, \vone^\T \vq = 1, \\
&& \frac{1}{n}\| n \vq - \vone\|_2^2 \leq \frac{\rho}{n} \big\}.
\eenrr
Here $\cQ_n$ is a set of $\vq$ and stands for a robust region. 
For MM-REx, $\tilde \Delta_n(\alpha)=\cQ_n(\alpha, +\infty).$ 
On the other hand, $\cQ_n(0,\rho)$ is a set of discrete distributions:
\benn
\big\{\vq:  \,\, \vq \in \R^n_{+}, \,\, \vone^\T \vq = 1,\,\, D_{\phi}(\vq\|\frac{1}{n}\vone) \leq \frac{\rho}{n} \big\}
\eenn
where $(1/n)\vone$ is the empirical distribution with $q_i=\frac{1}{n}$, $i=1,\ldots,n.$
In the following, we still use $D_{\phi}$ to measure the distance between two elements of $\cQ_n(\alpha, \rho).$
We define a quasi distributional robustness optimization (quasi DRO) problem, 
\ben\label{eq411}
\min_{h \in \cH} \max_{\vq \in \cQ_n(\alpha, \rho)} \,\, \vq^\T \hat{\rvr}_n(h). 
\een
This problem is an extension of distributional robustness optimization since $q_i$ can be negative.
The choices of $\lambda$ and $\rho$ determine the size of $\cQ_n$ and subsequently influence the robustness guarantees.
The problem (\ref{eq411}) unifies ERM, RO and MM-REx with different values of $\lambda$, since ERM and group DRO are special cases corresponding to $\cQ_n(-\frac{1}{n}, +\infty)$ and $\cQ_n(0, +\infty)$ respectively.
However, it is not easy to interpret the parameter $\lambda$, since $-\lambda$ is a lower bound of $q_i$.
Notice that, for ERM, $q_i \geq -\lambda = 1/n.$ Thus $\cQ_n(-\frac{1}{n}, +\infty)$ only contains one element $\frac{1}{n}\vone$ and can be rewritten as $\cQ_n(-\frac{1}{n}, 0).$ 
This motivates us to rewrite the robust region of ERM, RO and MM-REx. 
In the following, we shall prove that $\lambda$ and $\rho$ govern each other, and tuning $\lambda$ is equivalent to tuning $\rho.$

We start with an upper bound of the objective function in the problem (\ref{eq411}), which is independent to $\lambda.$
\begin{proposition}\label{proposition1}
Suppose the training data $\sD = \{\sD^e, e\in \cE_{tr}\}$ and the hypothetical model $h\in\cH$ are given. 
For any $-1/m \leq \alpha \leq +\infty$, 
\ben\label{eq412}
\max_{\vq \in \cQ_n(\alpha, \rho)} \vq^\T \hat{\rvr}_n(h) \leq \frac{1}{n} \vone^\T \hat{\rvr}_n(h) + \sqrt{\frac{\rho}{n} s_n^2(\hat{\rvr}_n(h))}.
\een
If $s_n(\hat{\rvr}_n(h))=0$, the equality in (\ref{eq412}) always holds. For $s_n(\hat{\rvr}_n(h))>0$, if $\alpha > C_{h, \sD, \rho}$, where 
\benn
C_{h, \sD, \rho}= -\frac{1}{n} + \frac{\sqrt{\rho}|\min_{e \in \cE_{tr}} \hat \rr_m(h,e) - \frac{1}{n}\vone^\T \hat \rvr_n(h)|}{n^{3/2} s_n(\hat{\rvr}_n(h)) },
\eenn
then the equality in (\ref{eq412})  holds.
\end{proposition}

Note that the inequality holds uniformly for $- 1/m \leq \alpha \leq +\infty.$
Thus, if the equality in (\ref{eq412}) can be achieved, the robust region $\cQ_n(\alpha, \rho)$ can deal with the robust optimization problem over $\cQ_n(+\infty, \rho).$
In other words, the RHS of (\ref{eq412}) can bound all possible linear combinations of the training risks $\vq^\T \hat{\rvr}_(h)$ for $\vq \in \cQ_n(\infty, \rho).$
The result shows that the equality can be achieved when $\alpha$ is sufficiently large. 

In fact, the scalar $C_{h, \sD, \rho}$ measures how hard it is to achieve universal robustness with respect to $\alpha.$  
If $C_{h, \sD, \rho}<0$, group DRO is equivalent to MM-REx.
According to the expression of $C_{h, \sD, \rho}$, we can understand three factors $n$, $\rho$ and $s_n(\hat \rvr_n(h))$ that influence the quasi DRO problem.
If $\rho$ is fixed, $C_{h, \sD, \rho}$ converges to zero in probability as $n$ tends to infinity. 
Hence more training environments benefits the robustness. 
Second, $C_{h, \sD, \rho}$ increases as $\rho$ increases.
It is trivial because $\rho$ determines the size of $\cQ_n(+\infty, \rho).$ 
For the third factor $s_n(\hat \rvr_n(h))$, the variance represents the diversity of the training domains, benefits the robust learning in general.

On the other hand, if $\alpha$ is fixed and $\rho$ is small enough such that
\benn
\rho \leq C'_{h,\sD, \alpha} := \frac{n (n \alpha +1)^2 s_n^2(\hat \rvr_n(h))}{2 \big(\min_{e\in\cE_{tr}} \hat \rr_m(h,e)- \frac{1}{n} \vone^\T \hat{\rvr}_n(h)\big)^2},
\eenn
then the equality in (\ref{eq412}) still holds. 
Thus $\alpha$ also governs $\rho.$
We further denote $\rho_{+} = D_\phi(\vq^*_{+}\|\frac{1}{n}\vone)$, where 
\benn
\vq^*_{+} = (1+n \alpha, 0, \ldots, 0)^\T - \alpha \vone 
\eenn
is a vertex of $\cQ_n(\alpha, +\infty)$ and $\rho_{+}$ is the largest distance between $\vq \in \cQ_n(\alpha, +\infty)$ and $\frac{1}{n}\vone.$

\begin{proposition}\label{proposition2}
Suppose that the training data $\sD$ and the hypothetical model $h\in \cH$ are given. If $\alpha$ is fixed and $\rho_{-} = C'_{h,\sD, \alpha}$, then we have 
\benr\label{eq413}
&& \frac{1}{n} \vone^\T \hat{\rvr}_n(h) + \sqrt{\frac{\rho_{-}}{n}} s_n(\hat{\rvr}_n(h)) \nn \\
&\leq& \max_{\vq \in \cQ_n(\alpha, +\infty)} \vq^\T \hat{\rvr}_n(h) \nn \\
&\leq& \frac{1}{n} \vone^\T \hat{\rvr}_n(h) + \sqrt{\frac{\rho_{+}}{n}} s_n(\hat{\rvr}_n(h)).
\eenr
\end{proposition}

Hence one can find $\rho^*\in [\rho_{-}, \rho_{+}]$ such that 
\benn
\max_{\vq \in \cQ_n(\alpha, +\infty)} \vq^\T \hat{\rvr}_n(h) 
= \frac{1}{n} \vone^\T \hat{\rvr}_n(h) + \sqrt{\frac{\rho^*}{n} s_n^2(\hat{\rvr}_n(h))}.
\eenn
It implies that in the problem (\ref{eq411}), tuning $\alpha$ is equivalent to tuning $\rho.$ 
In the following, we shall show that the LHS of (\ref{eq413}) can uniformly approximate MM-REx for any $\sD$ and $h$.

Next let's focus on $\rho_{-}$, the lower bound of $\rho^*$, and compare MM-REx and group DRO.
According to Proposition~\ref{proposition1} and~\ref{proposition2}, group DRO can bound all possible linear combinations of the training risks on $\cQ_n(0, C'_{h,\sD, 0}).$
At the same time, MM-REx can deal with the robust region $\cQ_n(0, C'_{h,\sD, \alpha}).$
From the view of group distributional robustness, the uncertainty region of MM-REx is is much larger than that of group DRO.
Hence the factor $(n\alpha + 1)^2$ in $\rho_{-} = C'_{h,\sD, \alpha}$ represents the potential benefit of risk extrapolation, which significantly enlarges $\cQ_n(0, C'_{h,\sD, 0}).$
In summary, MM-REx is more robust than group DRO by enlarging the uncertainty set $\cQ_n.$

\subsection{Risk variance penalization}\label{sec42}

According to the arguments in Section~\ref{sec41}, we know that: (1) the quasi DRO unifies ERM, group DRO and MM-REx; (2) the tuning parameters $\alpha$ and $\rho$ govern each other. 
These motivate us to consider the transformation from $\alpha$ to $\rho$, and to propose the following objective function:
\benn
\cL_{rvp}(h) = \frac{1}{n} \vone^\T \hat \rvr_n(h) + \lambda s_n(\hat \rvr_n(h)) 
\eenn
We call the method that $\hat h = \argmin_{h\in\cH} \cL_{rvp}(h)$ by Risk Variance Penalization (RVP). 
According to Proposition~\ref{proposition2}, RVP is equivalent to MM-REx. On the other hand, if we focus on the regularization term, $\cL_{RVP}$ is just a tiny change to $\cL_{V-REx}$. 
Therefore RVP connects V-REx to MM-REx and points out the relationship between the risk extrapolation and the variance penalization.

The proposed regularization is simple and easy to compute since it does not involve any maximization and gradient. 
Comparing to V-REx, the advantage of RVP is the interpretable tuning parameter $\lambda.$
By Proposition~\ref{proposition2}, we can rewrite $\lambda$ as $\sqrt{\rho/n}$, where
$\rho/n$ is the radius of the robust
region $\cQ_n(+\infty, \rho).$
On the other hand, we can also interpret $\lambda$ from the asymptotic view and give a theory-inspired tuning scheme. 
For simplicity, we assume $h$ is given and $m_e = \infty$, i.e. $\rr(h,e)$ is known.
Then 
\benn
\cL_{rvp}(h) = \frac{1}{n}\vone^\T \rvr_n(h) + \lambda s_n(\rvr_n(h))
\eenn
where $\rvr_n(h) = \big[ \cdots, \rr(h, e), \cdots \big]$ with $e\in\cE_{tr}.$
Next, we check the following probability
\benrr
&& \sP\big(\E_{e \sim Q}[\rr(h, e)] \leq \cL_{rvp}(h)\big) \\
&=& \sP\Big( \frac{\sqrt{n}(\frac{1}{n}\vone^\T \rvr_n(h) - \E_{e \sim Q}[\rr(h, e)])}{ s_n(\rvr_n(h))} \geq -\sqrt{n}\lambda \Big).
\eenrr
By the law of large numbers, $s^2_n(\rvr_n(h))$ converges in probability to
$\sigma^2_{\rr}:=\Var_{e \sim Q}[\rr(h, e)]$ as $n \rarrow +\infty.$
In addition, by the central limit theorem,  $\sqrt{n}(\frac{1}{n}\vone^\T \rvr_n(h) - \E_{e \sim Q}[\rr(h, e)])/\sigma_{\rr}$ converges in distribution to the standard normal distribution as  $n \rarrow +\infty.$
By the Slutsky's theorem, 
\benn
\sP\big(\E_{e \sim Q}[\rr(h, e)] \leq \cL_{rvp}(h)\big) \rarrow \Phi(\sqrt{n}\lambda)
\eenn
where $\Phi(\cdot)$ represents the cumulative density function of the standard normal distribution.
Therefore, we can determine $\lambda$ by $\Phi^{-1}(1-\gamma)/\sqrt{n}$ where $1-\gamma$ is the confidence level and $\Phi^{-1}(1-\gamma)$ stands for the $1-\gamma$ quantile of the standard normal distribution.


\subsection{Uniform Equivalence}\label{sec4}

Section~\ref{sec41} shows the equivalence between MM-REx and RVP when $h \in \cH$ and $\sD$ are given. 
In this section, we employ concentration
inequalities and Rademacher complexity to derive the uniform results on the equivalence between MM-REx and RVP.
According to the proof of Proposition~\ref{proposition2} in the Appendix, the second inequality in (\ref{eq413}) is trivial and always holds since $\cQ_n(\alpha, +\infty) \subset \cQ_n(+\infty, \rho_{+}).$ Thus,  (\ref{eq413}) can uniformly bound MM-REx from above without depending on the data $\sD.$
Next we refine the first inequality in (\ref{eq413}) to drive a uniform lower boundary of MM-REx, in which $\rho_{-}>0$ does not depend on $\sD.$

\begin{theorem}\label{theorem3}
Suppose that the loss function $\ell$ is bounded by $[0, M]$, $\alpha$ is a positive scalar and $m=\min_{e \in \cE_{tr}} m_e$ is sufficiently large such that $16M^2 <m\sigma_\rr^2$ where $\sigma^2_\rr=\Var_{e \sim Q}[\rr(h, e)].$
Let 
\ben\label{eq51}
\rho'_{-} = \frac{n(n\alpha+1)^2(1-\varepsilon)^2\sigma^2_{\rr}}{ M^2}
\een
where $\frac{4M}{\sqrt{m\sigma_{\rr}^2}}<\varepsilon<1.$
The following expansion holds:
\benn
\max_{\vq \in \cQ_n(\alpha, \rho'_{-})} \vq^\T \hat{\rvr}_n(h) 
= \frac{1}{n} \vone^\T \hat{\rvr}_n(h) + \sqrt{\frac{\rho'_{-}}{n}} s_n(\hat{\rvr}_n(h))
\eenn
with probability at least 
\benn
1-\exp\Big(-\frac{n \varepsilon^2 \sigma_{\rr}^2}{8M^2}\Big) - \exp\Big(-n \big(\frac{\varepsilon \sqrt{m\sigma_{\rr}^2}}{2M} - 2\big)^2\Big).
\eenn
\end{theorem}

Next we extend Theorem~\ref{theorem3} to a more general variant with respect to the family of the hypothetical models $\cH$.
Let $\{\rvz_1, \ldots, \rvz_n \}$ be a sample and $\epsilon_i \in \{+1, -1\}$ be i.i.d.
random signs independent of the sample. 
Denote $\cF$ as the collection of the bounded function $f=\ell(h, \cdot ): \rvz \rarrow [0,M]$ with $h \in \cH.$ 
The worst-case Rademacher complexity \cite{srebro2010smoothness} is given by
\benrr
\Re_n^{sup}(\cF) = \sup_{\rvz_1,\ldots, \rvz_n} \E\Big[\sup_{f \in \cF}\Big|\frac{1}{n}\sum_{i=1}^n \epsilon_i f(\rvz_i)\Big| \Big]
\eenrr
Using this definition, we prove that the expansion in Theorem~\ref{theorem3} holds uniformly for all functions in $\cF$ with high probability.

\begin{theorem}\label{theorem4}
There exists a universal constant $C$ such that if $1-\frac{\sqrt{2}}{2}<\varepsilon<1$ satisfies
\benrr
&& \big(\frac{1}{2} - (1-\varepsilon)^2\big)\sigma^2_\rr - \frac{1}{2}\E[\sigma^2_{f,e}] \\
&\geq& C \Big[\Re^{sup}_{nm} (\cF)^2 \log^3 (nm) + \frac{M^2}{nm}
(t + \log \log nm) \\
&& +M\sqrt{\frac{t}{2nm}}\Big],
\eenrr
where $\sigma^2_{f,e}$ is the variance of $\ell(h(\rvx), \rvy)$ under the domain $e.$
Then with probability at least $1 - 4\exp(-t)$,
\benrr
 \max_{\vq \in \cQ_n(\lambda, \rho'_{-})} \vq^\T \hat{\vr} 
 = \frac{1}{n} \vone^\T \hat{\rvr}_n(h) + \sqrt{\frac{\rho'_{-}}{n}} s_n(\hat{\rvr}_n(h))
\eenrr
for all $f = \ell(h, .)$ in $\cF.$
\end{theorem}

Next let's focus on $\rho_{-}$, the lower bound of $\rho^*$, and compare MM-REx and group DRO.
According to Proposition~\ref{proposition1} and~\ref{proposition2}, group DRO can bound all possible linear combinations of the training risks on $\cQ_n(0, C'(f,\sS, 0)).$
At the same time, MM-REx can deal with the robust region $\cQ_n(0, C'(f,\sS, \lambda)).$
From the view of distributional robustness, the uncertainty set of MM-REs is is much larger than that of group DRO.
Hence the factor $(m\lambda + 1)^2$ in $\rho_{-} = C'(f,\sS, \lambda)$ represents the potential benefit of risk extrapolation, which significantly enlarges $\cQ_n(0, C'(f,\sS, 0)).$
In summary, REx is more robust than RO by enlarging the uncertainty set $\cQ_n.$

\section{Experiments on PACS and VLCS}\label{sec:exp1}

This section provides two real data examples to show the robustness of RVP and verify the utility of our tuning scheme.
We employ \href{https://github.com/facebookresearch/DomainBed}{DomainBed} \citep{gulrajani2020search}, which is a testbed that aims to provide fair and realistic comparisons for domain generalization algorithms.
The experiments involve 2 datasets, PACS \citep{li2017deeper} and VLCS \citep{fang2013unbiased}, and 5 methods: ERM \citep{vapnik1992principles}, Group DRO \citep{sagawa2019distributionally},  IRM \citep{arjovsky2019invariant}, V-REx \citep{krueger2020out} and the proposed method RVP.
The model architectures and hyperparameters are the same as those in DomainBed.
Both PACS and VLCS consists of four domains. We split each domain into training and test subsets. In each experiment, we use three domains for training and all four domains for testing.
The evaluation metric is the OOD accuracy that is the worst-domain test accuracy over all four domains.
We report the best achievable OOD accuracy during 5000 iterations. 


\begin{table}
\centering
\begin{tabular}{|l|l|l|l|l|}
\hline
PACS & A & C & P & S\\ 
\hline
ERM & 0.8655 & 0.8356 & 0.9375 & 0.8108 \\
    & (0.0158) & (0.0117) & (0.0094) & (0.0237) \\
\hline
Group & 0.8716 & 0.8435 & 0.9403 & {\bf 0.8140} \\
 DRO  & (0.0152) & (0.0183) & (0.0076) & (0.0326) \\
\hline
RVP & {\bf 0.8736} & {\bf 0.8498} & {\bf 0.9412} & 0.8139 \\
$\lambda = 1.132$    & (0.0184) & (0.0178) & (0.0056) & (0.0227) \\
\hline
\hline
VLCS & C & L & S & V\\ 
\hline
ERM & 0.7609 & {\bf 0.6774} & 0.7324 & 0.7452 \\
    & (0.0144) & (0.0192) & (0.0192) & (0.0125) \\
\hline
Group & {\bf 0.7690} & 0.6753 & 0.7220 & 0.7372 \\
 DRO  & (0.0120) & (0.0219) & (0.0140) & (0.0167) \\
\hline
RVP & 0.7653 & 0.6747 & {\bf 0.7375} & {\bf 0.7464} \\
    & (0.0123) & (0.0196) & (0.0190) & (0.0083) \\
\hline
\end{tabular}
\caption{OOD accuracy of ERM, Group DRO and RVP.}
\label{tab:robust}
\vspace{-10pt}
\end{table}

We first compare RVP to ERM and Group DRO.
Notice that RVP is equivalent to MM-REx, which is an extension of Group DRO.
According to the theory-inspired scheme, we take the tuning parameter $\lambda$ to be $1.96/\sqrt{3}=1.132$ where $1.96$ is the $97.5\%$ quantile of the standard normal distribution and $3$ is the number of training domains.
We report the mean and standard deviation of OOD accuracy among 10 repetitions with the same random seeds.  
The results are presented in Table~\ref{tab:robust}.
On PACS, RVP is comparable to Group DRO and outperforms ERM.
On the other hand, the standard deviation of OOD accuracy is very large, which is almost ten times larger than the standard deviation of test accuracy on one domain (see \citet{gulrajani2020search}). This is reasonable because there are only four samples at the domain level.
On VLCS, we do not see the uniform improvement from ERM to RVP in terms of the averaged OOD accuracy. 
This observation is consistent with  \citet{ye2021out}, which shows that there may be no sufficient diversity among the domains in VLCS.

Next, we consider IRM and V-REx, which can discover invariant prediction under certain conditions. Both IRM and V-REx use large regularization parameter $\lambda$ to enforce the invariance over training domains. In the following, we take $\lambda$ to be $1$, $10$ and $100$. The results of IRM, V-REx and RVP are presented in Table~\ref{tab:invariant}.
One can find that RVP is comparable to IRM and V-REx and the theory-inspired tuning scheme performs well.

\begin{table}
\centering
\begin{tabular}{|l|l|l|l|l|}
\hline
PACS & A & C & P & S\\ 
\hline\hline
IRM & 0.8728 & 0.8376 & 0.9392 & 0.8123 \\
$\lambda = 1$ & (0.0203) & (0.0142) & (0.0083) & (0.0254) \\
\hline
IRM & 0.8166 & 0.7547 & 0.8850 & 0.7160 \\
$\lambda = 10$ & (0.0506) & (0.0563) & (0.0392
) & (0.0486) \\
\hline
IRM & 0.7264 & 0.7335 & 0.7539 & 0.6128 \\
$\lambda = 100$  & (0.0616) & (0.0594) & (0.1152) & (0.1001) \\
\hline
\hline
V-REx & 0.8723 & 0.8382 & {\bf 0.9414} & {\bf 0.8224} \\
$\lambda = 1$ & (0.0170) & (0.0134) & (0.0057) & (0.0136) \\
\hline
V-REx & 0.8647 & 0.8282 & 0.9398 & 0.8082 \\
$\lambda = 10$ & (0.0189) & (0.0124) & (0.0116) & (0.0190) \\
\hline
V-REx & 0.8418 & 0.8235 & 0.9292 & 0.7774 \\
$\lambda = 100$ & (0.0283) & (0.0126) & (0.0149) & (0.0285) \\
\hline
\hline
RVP & 0.8765 & 0.8307 & 0.9406 & 0.8178 \\
$\lambda = 1$ & (0.0168) & (0.0114) & (0.0090) & (0.0241) \\
\hline
RVP & 0.8736 & {\bf 0.8498} & 0.9412 & 0.8139 \\
$\lambda = 1.132$ & (0.0173) & (0.0159) & (0.0077) & (0.0226) \\
\hline
RVP & {\bf 0.8838} & 0.8414 & 0.9375 & 0.8170 \\
$\lambda = 10$ & (0.0147) & (0.0117) & (0.0083) & (0.0167) \\
\hline
RVP & 0.8440 & 0.7989 & 0.9263 & 0.7699 \\
$\lambda = 100$ & (0.0244) & (0.0308) & (0.0136) & (0.0347) \\
\hline
\end{tabular}
\caption{OOD accuracy of IRM, V-REx and RVP on PACS.}
\label{tab:invariant}
\vspace{-10pt}
\end{table}

{ \bf Discussions} For invariant prediction, the existing study employs multi-domain sourced data to learn an invariant predictor for training domains.
From a theory perspective, the key problem is to prove that the learned invariance can generalize from training domains to target domains.
One feasible route is causality. 
The causal relationship can guarantee the invariance over unseen target domains. Thus, the primary problem is how to discover the causal relationship from the invariance over training domains \citep{peters2016causal}. Another feasible route is distributional robustness, which weakens the definition of invariance. 
If we assume all domains are generated from a meta distribution, the generalization of robustness is directly embedding into the uncertainty set of the optimization problem.
The V-REx and RVP belong to the latter.
In this section, we have shown that RVP with a small $\lambda$ is comparable to Group DRO and can outperforms ERM. 
On the other hand, causal inference
can be seen as a special case of distributional robustness, where the uncertainty set covers all distributions
generated by do-interventions or shift-interventions
on a given causal model \cite{meinshausen2018causality,pearl2011graphical,spirtes2000causation}.
Note that the larger the regularization parameter $\lambda$, the larger the uncertainty set.
According to \citet{krueger2020out}, V-REx with a large $\lambda$ can discover an invariant predictor for Colored MNIST \citep{arjovsky2019invariant}.
In the following, we experimentally show that RVP can also discover an invariant predictor under certain conditions.

\section{Experiments on Colored MNIST}

The Colored MNIST dataset from \cite{arjovsky2019invariant} presents a binary classification task.
Its generating procedure is as follow:
First, label each image from the original MNIST by its digit. 
If the digit is from $0$ to $4$, then label the image with $\tilde y=0$; otherwise, label the image with $\tilde y=1.$ 
Second, obtain the final label $y$ by flipping $\tilde y$ with probability $P_\epsilon$. 
Third, generate the color label $z$ by flipping the final label $y$ with probability $P_i$, which represents the domain generating procedure.
Finally, color the image based on the color label $z$: red for $z = 1$ and green for $z = 0.$
It is easy to see that the digit determines the final label, and the final label determines the color.
Thus the causal (invariant) factor is the digit while the color is the spurious feature. 

In the following, we present two examples on Colored MNIST to show: (i) RVP can learn the invariant factor; (ii) The variance-based regularization can remove the variant factor.
We consider 9 domains corresponding to $P_i = i/10$, $i=1,\ldots,9$, and denote $\cP=\{P_1, \ldots, P_9\}.$
In addition, we take $P_\epsilon= 0.25$ and $0.5$. 
Here $P_\epsilon= 0.5$ is the baseline, which implies that the predictor only depends on the color factor.
Note that the OOD accuracy cannot be estimated by the performance on a single test domain \citep{ye2021out}. We record the test performance on each domain in $\cP$ and use the worst one to measure the OOD generalization.

\begin{figure}[ht]
\begin{center}
\centerline{\includegraphics[width=0.95\columnwidth]{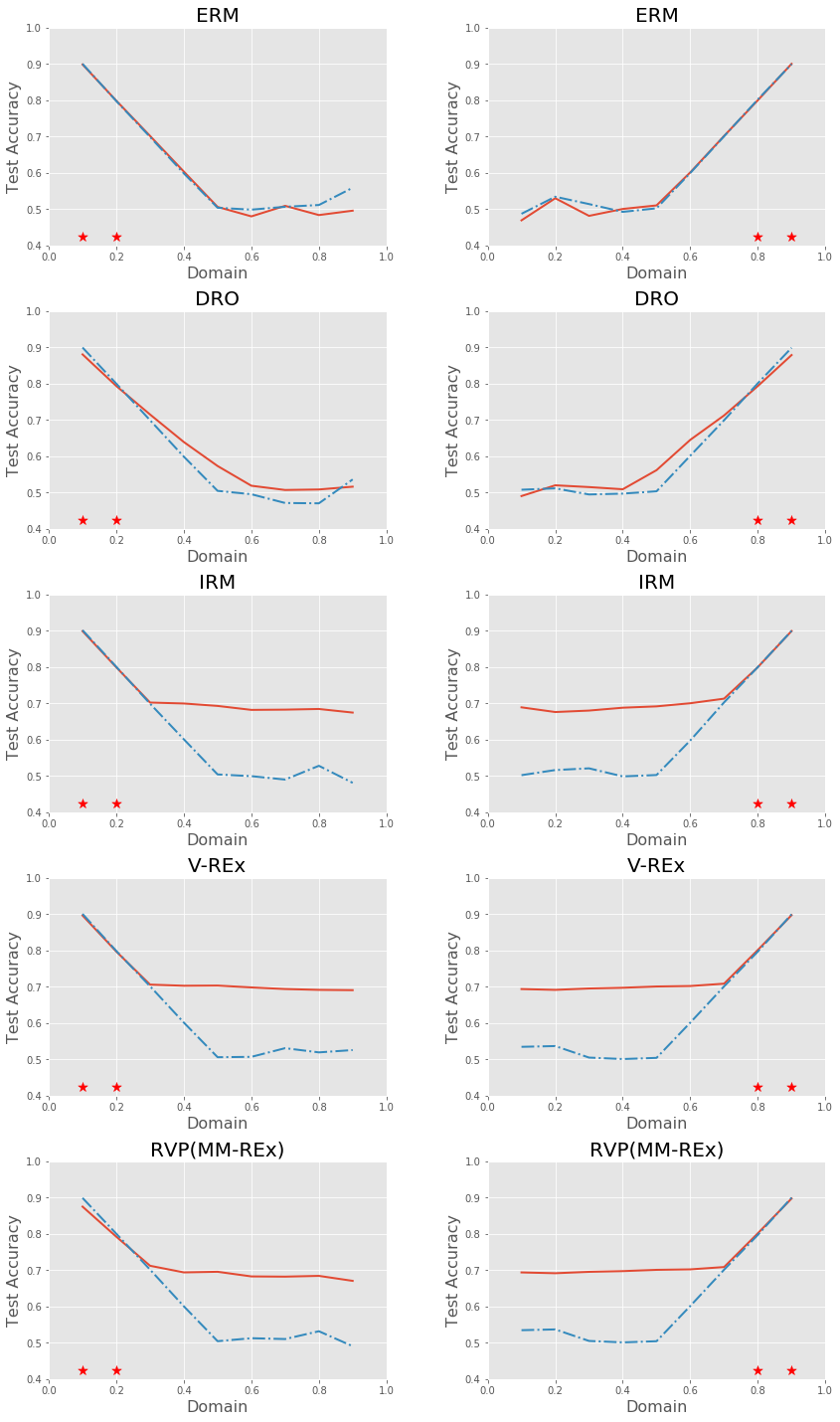}}
\caption{The best test accuracy before 500 epochs under two training domains. 
The red solid line and blue dotted line stand for $P_\epsilon = 0.25$ and $P_\epsilon = 0.5$ respectively.
The red stars represent the training domains.} 
\label{Fig1}
\end{center}
\vspace{-20pt}
\end{figure}

{\bf Example 1}. 
We consider $P_\epsilon = 0.25$ and $P_\epsilon =0.5$ and take two training environments from $\cP$:
\benn
\cP_{tr} = \{0.1, 0.2\}, \quad \text{and} \quad \cP'_{tr} = \{0.8, 0.9\}.
\eenn
The noise level $P_\epsilon= 0.5$ implies that all invariant features are removed from the data.
Since the test accuracy highly depends on $\lambda$ and other hyper-parameters, we report the best achievable accuracy during 500 epochs. 
The results are presented in Figure~\ref{Fig1}.
The accuracy under $P_\epsilon = 0.5$ is marked by the blue dotted line and the accuracy under $P_\epsilon = 0.25$ is marked by the red solid line. The stars represent the training environments.
The performance gap between $P_\epsilon= 0.25$ and $P_\epsilon = 0.5$ represents what learns from the invariant features (digit).
One can see that IRM, V-REx and RVP(MM-REx) can obtain good OOD performance via learning invariant features while ERM and Group DRO cannot.

\begin{figure}[ht]
\begin{center}
\centerline{\includegraphics[width=0.95\columnwidth]{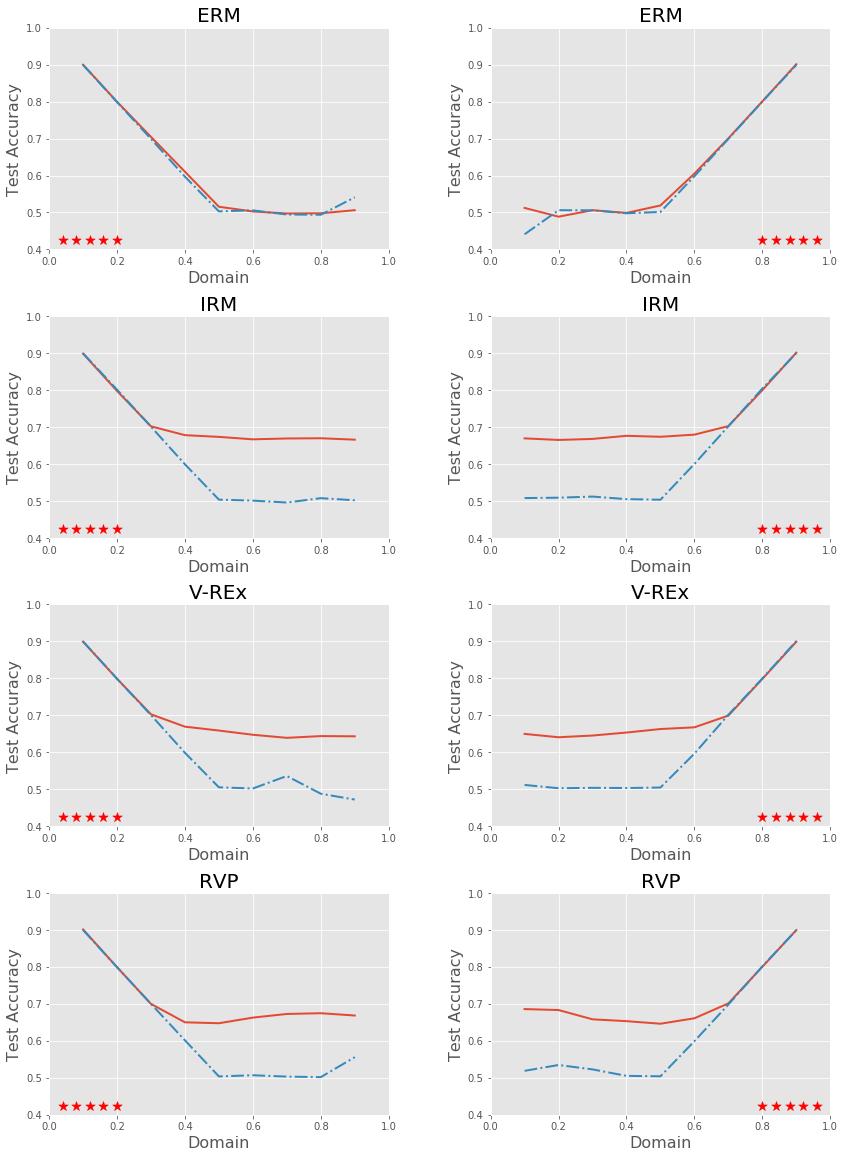}}
\vspace{-10pt}
\caption{The best test accuracy during 500 epochs under five training domains. The red solid line and blue dotted line stand for $P_\epsilon = 0.25$ and $P_\epsilon = 0.5$ respectively. The red stars represent the training domains.}
\label{Fig2}
\end{center}
\vspace{-15pt}
\end{figure}

Notice that MM-REx and RVP are exactly equivalent under two training domains. Thus, we also consider five training domains:
\benrr
\cP_{tr} &=& \{0.04, 0.08, 0.12, 0.16, 0.20\}, \\
\cP'_{tr} &=& \{0.80, 0.84, 0.88, 0.92, 0.96\}.
\eenrr
Here we report the best achievable accuracy of ERM, IRM, V-REx and RVP during 500 epochs. 
The results are presented in Figure~\ref{Fig2}.
One can see that IRM, REx and RVP can obtain good OOD performance by learning causal features.

\begin{figure}[htbp]
\begin{center}
\includegraphics[width=0.95\columnwidth]{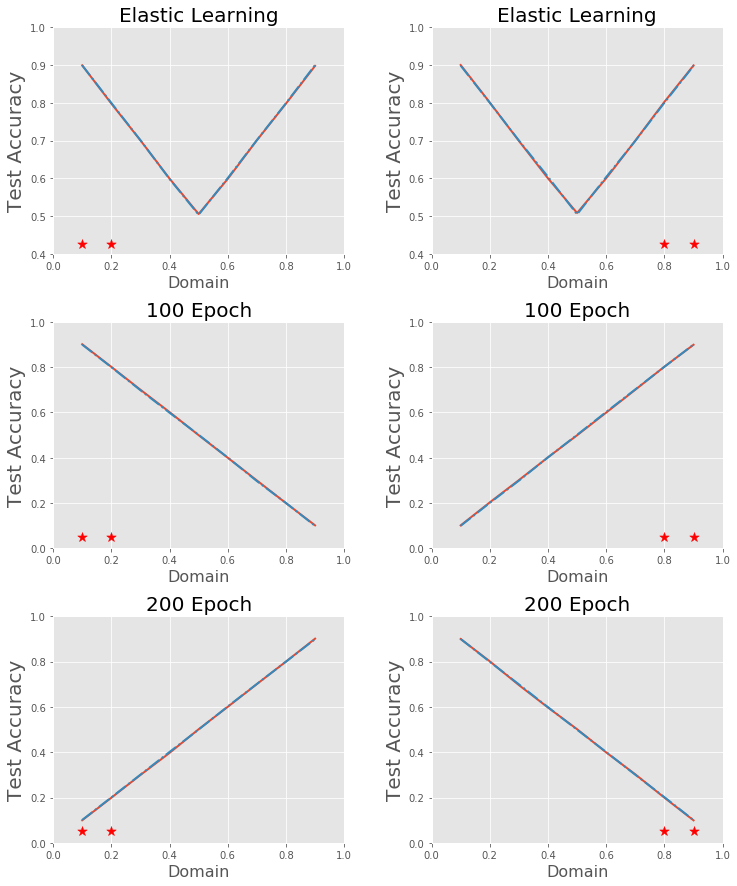}
\caption{The upper panels show the best test accuracy before 500 epoch. The middle and lower panels report the test accuracy of the learned model at Epoch 100 and Epoch 200. The red stars represent the training domains.}
\label{Fig3}
\end{center}
\vspace{-15pt}
\end{figure}

{\bf Example~2}. 
In this example, we design a learning scheme to check the utility of the variance-based regularization term in RVP. We assume $\lambda$ is infinity. Then RVP will find a solution such that the risk over training domains is invariant.
We design the following problem, denoted by Elastic Learning, 
\benn
\min_{h\in \cH}  \big\|\hat \rvr_n(h)\big\|_2^2, \,\,\text{subject to}\,\, \vone^\T \hat \rvr_n(h) \geq \lambda
\eenn
which enforces the equality of the risk across training domains.
The definition of $\hat \rvr_n(h)$ is given in Section~\ref{sec: preliminaries}.
The objective function is formulated into 
\benn
\cL_{EL}(h) = \big\|\hat \rvr_n(h)\big\|_2^2 - \lambda \vone^\T \hat \rvr_n(h).
\eenn
We select two training domains from $\cP$:
$\cP_{tr} = \{0.1, 0.2\}$ or $\cP_{tr} = \{0.8, 0.9\},$
and  $P_\epsilon = 0.25, 0.5$.
At the first $100$ epoch, $\lambda=1$ enforces $ \hat \rr_m(h, e) = 0.5$ for any $e \in \cP_{tr}.$  
After that, $\lambda$ is taken to be $10,000$. 
Thus the designed learning procedure will obtain a sequence of models such that $\hat \rr_m(h, e) \approx r$ for any $e \in \cP_{tr}$ with $r$ varying from 0.5 to 5000.
The best achievable accuracy is presented in Figure~\ref{Fig3}.

One can find that the best achievable accuracy under $P_\epsilon=0.25$ is almost the same to the best achievable accuracy under $P_\epsilon=0.5.$
This implies that the elastic learning ignores the invariant features (digit).
On the other hand, the elastic learning finds out two different prediction rules.
At the domains $P_1$ to $P_4$, the learned models use the positive correlation $y=z$ to predict the label $y$. 
This can be readily learned from the training data.
At the domains $P_6$ to $P_9$, the learned models use the negative correlation $y=-z$ to predict the label $y$. 
Furthermore, by extracting two models at Epoch 100 and Epoch 200, we find that the two models have learned two quite different prediction rules, which are $y = z$ and $y=1-z$ respectively.


\section{Conclusions}
This work investigates the variance-based regularization methods for domain generalization. We propose Risk Variance Penalization (RVP), which is robust to distribution shift across domains and has an interpretable regularization parameter. 
Furthermore, we experimentally show that under certain conditions, RVP can discover an invariant predictor.

\bibliography{main}
\bibliographystyle{icml2020}

\section{Appendix}

\subsection{Proof of Proposition \ref{proposition1}}

Since $\vone^\T \vq = 1$, we can decompose $\vq^\T \hat\rvr_n(h)$ into
\benrr
&& \vq^\T \hat{\rvr}_n(h) = \frac{1}{n} \vone^\T \hat{\rvr}_n(h) + \Big(\vq - \frac{1}{n}\vone\Big)^\T \hat{\rvr}_n(h)  \\
&=& \frac{1}{n} \vone^\T \hat{\rvr}_n(h) + \Big(\vq - \frac{1}{n}\vone \Big)^\T \Big(\hat{\rvr}_n(h) - \frac{1}{n} \vone \vone^\T \hat{\rvr}_n(h)\Big)
\eenrr
Let $\vv = \vq - \frac{1}{n}\vone.$
Thus the minimiax problem in (\ref{eq411}) is equivalent to
\benrr
&& \max_{\vv}\quad  \frac{1}{n} \vone^\T \hat{\rvr}_n(h) + \vv^\T \Big(\hat{\rvr}_n(h) - \frac{1}{n} \vone \vone^\T \hat{\rvr}_n(h) \Big)\\
&& \text{s.t.} \quad \vv + (\frac{1}{n}+\alpha)\vone \in \R^n_{+},\,\, \vone^\T \vv = 0,\,\, \|\vv\|_2^2 \leq \frac{\rho}{n^2}.  \nn
\eenrr
By the Cauchy-Schwarz inequality, 
\benr\label{A1}
&& \vv^\T \Big(\hat{\rvr}_n(h) - \frac{1}{n} \vone \vone^\T \hat{\rvr}_n(h)\Big) \nn \\
&\leq& \|\vv\|_2 \Big\| \hat{\rvr}_n(h) - \frac{1}{n} \vone \vone^\T \hat{\rvr}_n(h) \Big\|_2 \nn\\
&\leq& \sqrt{\frac{\rho}{n^2}} \sqrt{n} s_n(\hat{\rvr}_n(h)).
\eenr
The second inequality holds since 
\benn
s_n^2(\hat{\rvr}_n(h)) = \frac{1}{n} \big\|\hat{\rvr}_n(h) - \frac{1}{n} \vone \vone^\T \hat{\rvr}_n(h) \big\|_2^2.
\eenn
The equality in (\ref{A1}) is attained if and only if the vector $\vv$ satisfy: (i) $\vv$ and $\hat{\rvr}_n(h) - \frac{1}{n} \vone \vone^\T \hat{\rvr}_n(h)$ are in the same direction; (ii) $\|\vv\|_2^2 = \frac{\rho}{n^2}.$ This implies that the $i$-th element of $\vv$ should be
\benrr
v_i &=& \sqrt{\frac{\rho}{n^2}}\frac{\hat \rr_m(h, e_i) - \frac{1}{n}\vone^\T \hat{\rvr}_n(h)}{ \|\hat{\rvr}_n(h) - \frac{1}{n} \vone \vone^\T \hat{\rvr}_n(h)\|_2 } \\
&=& \sqrt{\frac{\rho}{n^2}}\frac{\hat \rr_m(h, e_i) - \frac{1}{n}\vone^\T \hat{\rvr}_n(h)}{ \sqrt{n s_n^2(\hat{\rvr}_n(h))} },
\eenrr
where $e_i \in \cE_{tr}$ is the $i$-th trianing domain.
The only constraint here is $v_i \geq -\frac{1}{n}-\alpha$ which holds if and only if for any $1\leq i \leq n$,
\benn
\alpha \geq -\frac{1}{n} -  \sqrt{\frac{\rho}{n^2}}\frac{\hat \rr_m(h, e_i) - \frac{1}{n}\vone^\T \hat{\rvr}_n(h)}{ \sqrt{n s_n^2(\hat{\rvr}_n(h))}}.
\eenn
Hence the proof of Proposition~1 is finished.

\bbox

\subsection{Proof of Proposition \ref{proposition2}}

 Since $\rho_{+}$ is the largest distance between  $\frac{1}{n}\vone$ and $\vq \in \cQ_n(\alpha, +\infty)$, then $\cQ_m(+\infty, \rho_{+})$ covers $\cQ_m(\alpha, +\infty).$
Hence the second inequality in (\ref{eq413}) is trivial.
According to the proof of Proposition~\ref{proposition1}, we have
\benrr
\vq_{-}^* &=& \argmax_{\vq \in \cQ_n(+\infty, \rho_{-})} \vq^\T \hat{\rvr}_n(h) \\
&=& \frac{1}{n} \vone +  \sqrt{\frac{\rho_{-}}{n^2}}\frac{\hat{\rvr}_n(h) - \frac{1}{n}\vone^\T \hat{\rvr}_n(h)}{ \sqrt{n s_n^2(\hat{\rvr}_n(h))}}.
\eenrr
By the definition of $\rho_{-}$, $\vq^*_{-}$ belongs to $\cQ_n(\alpha, +\infty).$ Hence the proof of Proposition~2 is finished.

\bbox

\subsection{Proof of Theorem \ref{theorem3}}

{\bf Proof}: Note that $\hat \rr_m(h, e) \in [0, M]$ for any $h \in \cH$ and $e \in \cE_{tr}.$ Thus for any data $\sD$,
\benrr
C'_{h,\sD, \alpha} &=& \frac{n (n \alpha +1)^2 s_n^2(\hat{\rvr}_n(h))}{2 \big(\min_{e \in \cE_{tr}} \hat \rr_m(h,e)- \frac{1}{n} \vone^\T \hat{\rvr}_n(h)\big)^2} \\
&\geq& \frac{n (n \alpha +1)^2 s_n^2(\hat{\rvr}_n(h))}{M^2}.
\eenrr
Hence, to satisfying $\rho'_{-} \leq C'_{h,\sD, \alpha}$, it suffices to show that
\benrr
&& \rho'_{-} \leq \frac{n (n \alpha +1)^2 s_n^2(\hat{\rvr}_n(h))}{ M^2} \\
& \Leftrightarrow &  s_n(\hat{\rvr}_n(h)) \geq (1-\varepsilon)\sigma_{\rr}.
\eenrr
Define two events
\benrr
\cE_{n,1} &=& \Big\{ s_n(\rvr_n(h)) \geq (1- \frac{\varepsilon}{2}) \sigma_{\rr} \Big\} \\
\mathcal{E}_{n,2} &=& \Big\{ \big|s_n(\hat{\rvr}_n(h)) - s_n(\rvr_n(h))\big| \leq \frac{\varepsilon}{2}\sigma_{\rr} \Big\}, 
\eenrr
where $\rvr_n(h) = [\cdots,\rr(h,e),\cdots]$ with $e \in \cE_{tr}.$
Next we show that these two events hold with high probability. 

We start with the event $\mathcal{E}_{n,1}.$ 
According to Theorem 10 in \citet{maurer2009empirical} and Lemma 11 in \citet{duchi2019variance}, for $n \geq 2$,
\benrr
&& P\big(s_n(\rvr_n(h)) \geq  \sigma_\rr + t \big) \vee P\big(s_n(\rvr_n(h)) \leq  \sigma - t\big) \\
&\leq& \exp(-\frac{n t^2}{2 M^2}).
\eenrr
Let $t = \varepsilon \sigma_{\rr}/2.$ Then 
\benn
P(\cE_{n,1}) \geq 1- \exp(\frac{n \varepsilon^2 \sigma_{\rr}^2}{8 M^2}).
\eenn

For the event $\cE_{n,2}$, 
\benn
s_n(\rvr) = \frac{1}{\sqrt n} \Big\| \rvr - \frac{1}{n} \vone \vone^\T \rvr \Big\|_2
\eenn
is a convex and Lipschitz continuous function of $\rvr$ with respect to the $L_2$-norm over $\R^n.$ Then we have
\benn
\big|s_n(\hat{\rvr}_n(h)) - s_n(\rvr_n(h))\big| \leq \frac{1}{\sqrt n} \big\| \hat{\rvr}_n(h) -  \rvr_n(h) \big\|.
\eenn
Notice that the loss function $\ell(h, \rvz)$ is bounded and, the
the $i$-th element of $\hat{\rvr}_n(h)$ is the sample average approximation of the $i$-th element of $\rvr_n(h).$
Thus, by the concentration of the norm of sub-Gaussian random vectors \citep{vershynin2019high}, 
\benrr
&& \big|s_n(\hat{\rvr}_n{h}) - s_n(\rvr_n(h))\big| \\
&\geq& \frac{1}{m \sqrt n}\Big(2M\sqrt{m n} + M \sqrt{m \log(\frac{1}{\delta})}\Big),
\eenrr
with probability at most $\delta.$
Thus the event $\mathcal{E}_{n,2}$ holds with probability with probability at least
\benn
1- \exp\Big(-n \big(\frac{\varepsilon \sqrt{m\sigma_{\rr}^2}}{2M} - 2\big)^2\Big).
\eenn
Combining the results of $\mathcal{E}_{n,1}$ and $\mathcal{E}_{n,2}$, we know the expansion 
\benn
\max_{\vq \in \cQ_n(\alpha, \rho'_{-})} \vq^\T \hat{\rvr}_n(h) 
= \frac{1}{n} \vone^\T \hat{\rvr}_n(h) + \sqrt{\frac{\rho'_{-}}{n}} s_n(\hat{\rvr}_n(h))
\eenn
holds with probability at least
\benn
1-\exp\Big(-\frac{n \varepsilon^2 \sigma_{\rr}^2}{8M^2}\Big) - \exp\Big(-n \big(\frac{\varepsilon \sqrt{m\sigma_{\rr}^2}}{2M} - 2\big)^2\Big).
\eenn
Hence the proof of Theorem~\ref{theorem3} is finished.

\bbox

\subsection{Proof of Theorem \ref{theorem4}}

In this proof, we are going to give a uniform lower bound of $s_n^2(\rvr_n(h))$ that holds with high probability. 
Let the total variance of the loss function over all training data points be $s_{nm}^2(\rvr_{nm}(h))$, where
\benrr
s_{nm}^2(\rvr) &=& \frac{1}{nm}\big\|(\mI_{nm} - \frac{1}{nm}\vone \vone^\T) \rvr\big\|_2^2, \\
\rvr_{nm}(h) &=& [\cdots, \ell(h(\rvx_i^e), \rvy_i^e),\cdots], 
\eenrr
for $e \in \cE_{tr}$ and $i \in \{1,\ldots, m\}.$
Furthermore, we denote the in-domain variance of the loss function as $s_m^2(\rvr_m(h, e))$, where
\benrr
s_m^2(\rvr) &=& \frac{1}{m} \big\|(\mI_{m} - \frac{1}{m}\vone \vone^\T) \rvr\big\|_2^2, \\
\rvr_{m}(h,e) &=& [\cdots, \ell(h(\rvx_i^e), \rvy_i^e),\cdots], 
\eenrr
for $i \in \{1,\ldots, m\}.$
According to the decomposition of the total variance, we are ready to state 
\benn
s_n^2(\hat{\rvr}_n(h)) = s_{nm}^2(\rvr_{nm}(h)) - \frac{1}{n}\sum_{e \in \cE_{tr}} s_m^2(\rvr_m(h, e)).
\eenn
In this following, we are going to give a uniform lower bound of $s_{nm}^2(\rvr_{nm}(h))$ and a uniform upper bound of $s_m^2(\rvr_m(h, e)).$

We start with $s_{nm}^2(\rvr_{nm}(h)).$ Notice that, for any $nm$-length random vector $\rvr$,
\benrr
s_{nm}^2(\rvr) &=& \frac{1}{nm}\big\|(\mI_{nm} - \frac{1}{nm}\vone \vone^\T) \rvr\big\|_2^2 \\
&=& \frac{1}{nm} \big\| (\mI_{nm} - \frac{1}{nm} \vone \vone^\T ) (\rvr - \E[\rvr] ) \big\|_2^2 \\
&=& s_{nm}^2(\rvr - \E[\rvr]).
\eenrr
Then,
\benn
s_{nm}^2(\rvr) = \frac{1}{nm} \big\| \rvr - \E[\rvr] \big\|_2^2 - \big(\frac{1}{nm} \vone^\T (\rvr - \E[\rvr])\big)^2.
\eenn
Here $\E$ stands for the expectation with respect to the convolution distribution $\int P^e d Q(e).$
According to the Lemma~12 in \cite{duchi2019variance}, we know that
Then, with probability at least $1 - \exp(-t)$, for every $f \in \cF$, there exists a universal constant $C$ such that
\benr\label{A2}
\sigma^2_{f} &\leq& 2 \frac{1}{nm} \big\| \rvr_{nm}(h) - \E[\rvr_{nm}(h)] \big\|_2^2 \nn \\
&& + C \Big[\Re^{sup}_{nm} (\cF)^2 \log^3 (nm) \nn \\
&& + \frac{M^2}{nm}
(t + \log \log nm)\Big], 
\eenr
where $\sigma^2_{f}$ stands for the variance of $\ell(h(\rvx),\ry)$ with respect to the  convolution distribution $\int P^e d Q(e).$
For the upper bound of $\big(\frac{1}{nm} \vone^\T (\rvr_{nm}(h) - \E[\rvr_{nm}(h)])\big)^2$, we refer to the Lemma 14 in \cite{duchi2019variance}, which is a variant of Talagrand's inequality due to \citet{bousquet2002bennett,bousquet2003concentration}. 
(See also \cite{bartlett2005local}.)
Then, we have that, with probability at least $1- \exp(-t)$, 
\benrr
&& \sup_{f \in \cF} \frac{1}{nm} \vone^\T (\rvr_{nm}(h) - \E[\rvr_{nm}(h)]) \\
&\leq& \inf_{a>0} \Big\{2(1+a) \E\big[\Re_{nm}(\cF)\big] + M \sqrt{\frac{2t}{nm}}\\
&& + M \frac{t}{nm}\big(\frac{1}{3} + \frac{1}{a}\big)\Big\}.
\eenrr
The same statement holds with replacing the left-hand side of the
inequalities by 
$
\sup_{f \in \cF} \frac{1}{nm} \vone^\T (\E[\rvr_{nm}(h)] - \rvr_{nm}(h)).
$
By taking $a = 1/2$, we have with probability at least $1 - 2 \exp(-t)$,
\benr\label{A3}
&& \Big|\frac{1}{nm} \vone^\T (\rvr_{nm}(h) - \E[\rvr_{nm}(h)])\Big| \nn \\
&\leq& 3\E\big[\Re_{nm}(\cF)\big] + 2M \sqrt{\frac{2t}{nm}}
\eenr
holds for all $f\in \cF.$

Next we deal with the upper bound for $s_m^2(\rvr_m(h, e)).$ Similar to the arguments of $s_{nm}^2(\rvr)$, we know that 
\benrr
s_m^2(\rvr) &=& s_m^2\big(\rvr - \E_e[\rvr]\big) \\
&=& \frac{1}{m} \big\| \rvr - \E_e[\rvr] \big\|_2^2 - \big(\frac{1}{m} \vone^\T (\rvr - \E_e[\rvr])\big)^2.
\eenrr
Here $\E_e$ stands for the expectation with respect to the distribution $P^e$ and $\rvr$ is a $m$-length random vector.
It is easy to see that 
\ben\label{A4}
\frac{1}{n}\sum_{e\in\cE_{tr}} \big(\frac{1}{m} \vone^\T (\rvr_m(h, e) - \E_e[\rvr_m(h, e)])\big)^2 \geq 0.
\een
Notice that,
\benrr
&& \frac{1}{n} \sum_{e\in\cE_{tr}} \frac{1}{m} \big\| \rvr_m(h, e) - \E_e[\rvr_m(h, e)] \big\|_2^2 \\
&=& \frac{1}{nm} \sum_{e\in\cE_{tr}} \sum_{i=1}^m \big( \ell(h(\rvx_i^e), \rvy_i^e) - \E_e[\ell(h(\rvx_i^e), \rvy_i^e)] \big)^2. 
\eenrr
Then we have, with probability at least $1 - \exp(-t)$, 
\benr\label{A5}
&& \frac{1}{n} \sum_{e\in\cE_{tr}} \frac{1}{m} \big\| \rvr_m(h, e) - \E_e[\rvr_m(h, e)] \big\|_2^2 \nn \\
&\leq& \E[\sigma_{f,e}^2] + 2 M \Re_{nm}(\cF) + M\sqrt{\frac{t}{2nm}},
\eenr
where $\sigma_{f,e}^2$ is the variance of $\ell(h(\rvx), \rvy)$ with respect to the distribution $P^e$ and $\E[\sigma_{f,e}^2]$ stands for the in-domain variance of  $\ell(h(\rvx), \rvy).$

Combining the results of (\ref{A2}) to (\ref{A5}), then Theorem~\ref{theorem4} is proved.

\bbox

\end{document}